# CliqueParcel: An Approach For Batching LLM Prompts That Jointly Optimizes Efficiency And Faithfulness

Jiayi Liu*
liu2861@purdue.edu
Department of Computer Science
Purdue University
West Lafayette, Indiana, United States

Tinghan Yang
yang1683@purdue.edu
Department of Computer Science
Purdue University
West Lafayette, Indiana, United States

Jennifer Neville*
neville@purdue.edu
Microsoft Research
Redmond, Washington, United States

## ABSTRACT

Large language models (LLMs) have become pivotal in recent research. However, during the inference process, LLMs still require substantial resources. In this paper, we propose CliqueParcel, a method designed to improve the efficiency of LLMs via prompt batching. Existing strategies to optimize inference efficiency often compromise on output quality, leading to a *"discounted output"* problem. This issue might result in reduced accuracy or outputs that are less detailed. CliqueParcel is our answer to this challenge. While ensuring accuracy and minimizing deviations from the original outputs (i.e., faithfulness), our method significantly improves efficiency during inference.

To lay the groundwork, we first redefine efficiency measurements by excluding the reduction in running time due to shorter lengths. Then, we provide a comprehensive trade-off between efficiency and faithfulness to clarify the nature of the *"discounted output"* problem. Within the CliqueParcel framework, we suggest multiple batching sub-methods and discuss the specific scenarios in which they can be applied. During evaluation, CliqueParcel is tested on eight widely recognized datasets, which can be classified into three types: reading comprehension, open-source question-answering, and reasoning. Our experiments explore the performance of CliqueParcel, including efficiency, faithfulness, and the trade-off between them. This work provides novel insights into inference efficiency and demonstrates promising performance.

## 1 INTRODUCTION

Pre-trained large language models (LLMs) have achieved significant success in many natural language tasks[4, 25], including both zero-shot and few-shot setups[19, 44]. ChatGPT, as one of the most typical LLM tools, has reached over 100 million users and costs over 3 million USD per month to run[11]. While LLMs have achieved state-of-the-art (SOTA) performance compared to many other models, a question arises: How can the efficiency of LLMs be improved?

To address this question, recent research has explored optimization techniques in both the pre-training phase of LLMs and their subsequent use for specific tasks. The optimization efforts during the pre-training phase aim to enhance the efficiency of LLMs, which we refer to as the *pre-LLM* process[12, 13, 20, 37, 43], enabling faster training and reduced resource consumption. Other methods improve LLM efficiency through a *post-LLM* process, focusing on the inference phase. Some of these methods optimize the LLM models, such as quantization[9, 40, 42] and pruning[21, 26], while others choose to optimize prompts[6, 38, 41].

However, in *post-LLM* optimization, most of them achieve higher efficiency by introducing significant precision loss[9, 40, 42], or with an insufficient answer compared to the original one[6, 38, 41]. We refer to the phenomena above as the *"discounted output"* problem and model it as a discussion of faithfulness. This paper tackles the challenge of improving efficiency while maintaining faithfulness.

To solve this problem, we propose CliqueParcel, a method for efficiency-faithfulness trade-off optimization from the *post-LLM* aspect. CliqueParcel batches prompts across different users to facilitate an efficient computing process across similar domains. Despite its efficiency, CliqueParcel successfully generates correct answers to questions, ensuring the accuracy of LLM answers. First, we propose a small pre-model before LLM to classify the prompts according to their clique domain. Next, we batch the prompts with the same concept domain together and input them into LLM. Finally, we propose a disassembly step from the LLM output and dispatch corresponding answers to users. We demonstrate that CliqueParcel significantly improves efficiency compared to traditional prompt flow and other baselines while maintaining faithfulness.

Comprehensive empirical evaluations are conducted on eight widely recognized question-answering datasets: SQuAD, HotpotQA, TREC, CSQA, GSM8K, MATH, ANLI, and MMLU. Our findings clearly illustrate that CliqueParcel consistently outperforms standard prompting techniques and other baselines. Notably, CliqueParcel not only enhances the overall efficiency of the question-answering process but also ensures the faithfulness and accuracy of the answers provided.

Our main contributions are:

(1) We introduce the novel *"discounted output"* problem in the post-LLM process and propose the definition of efficiency and faithfulness to address it. Additionally, we introduce

(1) a new efficiency metric designed to mitigate the impacts of extraneous factors, thereby providing a more accurate measure of inference efficiency.
(2) Our framework, CliqueParcel, improves the efficiency of the post-LLM problem compared to existing baselines. We introduce multiple methods for the batching process of CliqueParcel, and propose feasible solutions for applying different batching methods in different scenarios.
(3) Through comprehensive experiments and analysis, we show that CliqueParcel effectively balances algorithmic efficiency and faithfulness.

## 2 RELATED WORK
### 2.1 Faithfulness in Large Language Model

Large language models (LLM) aims to estimate the probability over text. The main difference between large language models and other language models lies in the model size. Large language models are pre-trained and fine-tuned, which is usually static to users. LLMs are widely used in context learning, including few-shot learning and zero-shot learning[19].

With the development of LLMs, there is an increasing focus on addressing faithfulness issues in LLMs' generated outputs[1, 30, 33]. In the realm of natural language processing (NLP) tasks, faithfulness has gained prominence as an essential aspect for evaluating the quality of generated answers[2, 16, 27]. Faithfulness, in the context of NLP, refers to the extent to which a generated response or translation accurately reflects the information present in the source text or adheres to a given reference.

One popular criterion for assessing faithfulness is to use metrics like Bleu[23] or Rouge[8, 10]. These metrics have been adapted from machine translation and text summarization tasks to evaluate the faithfulness of generated responses across various NLP models and tasks. They provide quantitative measures that assess the overlap between the generated output and reference text, offering insights into the faithfulness of the generated content.

### 2.2 Efficiency of post-training large language models

There exist numerous methodologies aiming at enhancing the efficiency of large language models post-training. Current literature in this domain can be broadly categorized into two main areas: model compression and prompt compression. Within the realm of model compression, three prominent techniques have garnered significant attention: quantization, distillation, and pruning. Quantization is a widely employed approach in the compression of post-training large language models[9, 40, 42]. By reducing the bit precision, quantization methods necessitate lower memory utilization, thus facilitating faster computations and, consequently, higher efficiency. Distillation methods distill the knowledge from larger models and embed it into smaller models to approach model compression[5, 15, 17]. Another noteworthy method for model compression is pruning[21, 26]. Pruning achieves reduced complexity by eliminating relatively less impactful model components, effectively reducing the model's size.

On the other hand, prompt-based techniques contribute to efficiency enhancement by modifying the prompts used in interactions with large language models[6, 38, 41]. These approaches seek to optimize the prompts to yield improved results in terms of efficiency. Xu et al. attach a prompt to get less accurate outputs, which helps in finding the efficiency-accuracy trade-off[38]. Yin et al. focus on finding the task definition to shorten prompts[41]. Cheng et al. use a batching method to improve efficiency. However, all those prompt-based techniques has a common problem: they improve their efficiency mainly due to an incompleteness of the outputs.

## 3 APPROACH
### 3.1 Motivation

The efficiency discussion of LLM has not been fully discussed yet. Since LLM takes billions of request per seconds, improving the efficiency of LLM and reducing the memory/card usage is important. Most current LLM based models are bert-based, which have a maximum sequence length due to memory constraints. By merging sentences, we can utilize this maximum sequence length more effectively. For example, instead of processing two sentences of length 50 separately (and wasting the remaining space in the 512-length sequence), we can merge them into a single sequence and process them together.

### 3.2 Problem definition

We start by formalizing the problem as an assignment problem. Suppose we have $m$ prompts $p_1, p_2, ... p_m$ from $n$ users $N_1, N_2, ... N_n$ with $d : \mathbb{P}^m \to \mathbb{N}^n$, while each prompt $p_i$ has a corresponding response answer $a_i = LLM(p_i)$, generalize as $\mathbb{A} = LLM(\mathbb{P})$. Ideally, we batch $\mathbb{P}' \in \mathbb{P}$ prompts to get a new prompt $P'$, name it $P' = f(\mathbb{P}')$. Then, we will get the answer as $A' = LLM(P')$.

We further make the assumption that there exists a function $g(\cdot)$ to batch the answers with $A' = g(\mathbb{A}')$. Then the equation would be

$$\begin{aligned} A' &= LLM(P') = LLM(f(\mathbb{P}')) \\ &= g(\mathbb{A}') = g(LLM(\mathbb{P}')) \end{aligned} \quad (1)$$

Here, we have $\mathbb{A}' = g^{-1}(LLM(f(\mathbb{P}')))$. The main challenge in this problem lies in three aspects: First, getting the criteria for batching prompts. Second, how to design the batch function $f(\cdot)$. Third, how to get the dispatch function $g^{-1}(\cdot)$.

To solve the challenges, we make a simplification to the problem stated above. First, we make the assumption that each prompt could be classified as a clique domain $h : \mathbb{P}^m \to \mathbb{C}^c$, corresponding to $C_1, C_2, ... C_c$ cliques. Second, we group the prompts based on their clique, naming it $\mathbb{P}'_k$, which satisfies $h(p_i) = C_k, \forall p_i \in \mathbb{P}'_k$. We call the group process as clique function $H(\cdot)$, with $H(\mathbb{P}) = \mathbb{P}'_1, \mathbb{P}'_2, ..., \mathbb{P}'_k, ...$. Third, we get the dispatch function $g^{-1}(\cdot)$ with a small trick, by pushing parts of the clues in $f(\cdot)$ design.

In the following sections, we will introduce our definition of efficiency and faithfulness, and why they exist. Our goal is to find a trade-off between efficiency and faithfulness.

*3.2.1 Faithfulness.* Previous methods have often overlooked the significance of answer faithfulness in discussions about LLM efficiency. Our goal is to establish a measure of answer faithfulness that closely aligns with the original output while maintaining a reasonable level of accuracy. To achieve this goal, we introduce the concept of faithfulness and define it based on three key aspects:

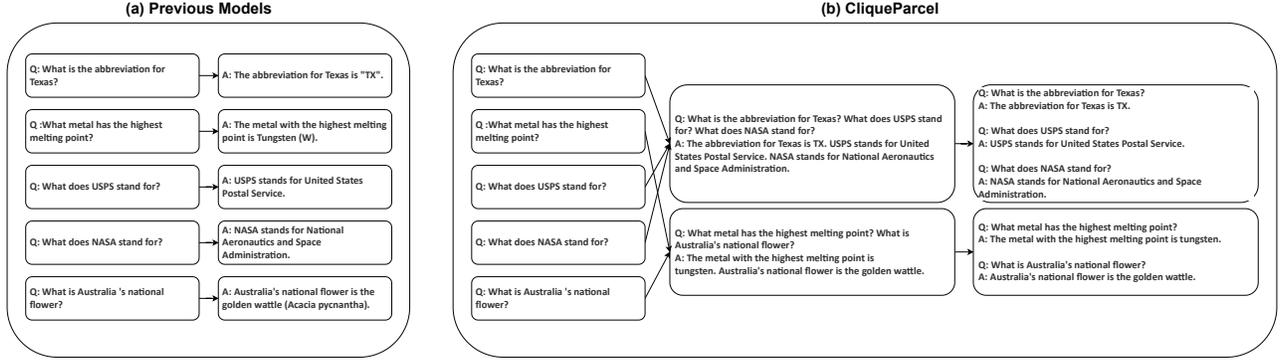

**Figure 1: Example inputs and outputs of GPT-3.5 with (a) standard prompt flow one by one, (b) CliqueParcel . Different from standard prompt flow, CliqueParcel batches the prompts with the same clique domain and put them into LLMs, then dispatches the corresponding answers to users.**

(1) We assess faithfulness based on the semantic cosine similarity between the original answer $a_i \in \mathbb{A}$ and the generated answer $a'_i \in \mathbb{A}'$. The essence of faithfulness is to ensure that the answers generated by our method closely mirror the original responses. To achieve this, we employ a metric that quantifies the cosine similarity between $a_i$ and $a'_i$.

(2) We utilize a Bleu and Rouge score to measure the quality of generated answer $a'_i$ towards the original answer $a_i$. Our guiding principle in this context is that any compression methods employed should continue to deliver *complete* answers rather than *'discounted'* ones. This approach aligns with the user's potential interest in accessing more comprehensive information. While semantic distance effectively measures the similarity between $a_i$ and $a'_i$, the inclusion of a weighted coefficient is crucial. It helps gauge whether the answer remains as complete as it was initially.

(3) We also add accuracy to be parts of the faithfulness definition. Adding accuracy as a component of the faithfulness measurement ensures that the generated answers not only resemble the original ones but also provide accurate information, meeting the user's expectations for reliable responses.

In essence, we introduce faithfulness to encompass both semantic similarity and factual correctness. It represents our commitment to delivering responses that not only meet the user's information needs but also uphold the highest standards of reliability.

In our pursuit of faithfulness, our primary objective is to maximize the faithfulness $D_h$. We first consider $d(\mathbb{A}', \mathbb{A})$, which is the faithfulness between the generated answers $\mathbb{A}'$ and the original answers $\mathbb{A}$. We meticulously measure this faithfulness, taking into account the three key factors mentioned earlier:

$$d(\mathbb{A}', \mathbb{A}) = \sum_{a_i \in \mathbb{A}, a'_i \in \mathbb{A}'} cos(a'_i, a_i) \cdot (BLEU(a'_i, a_i) + ROUGE(a'_i, a_i)) \cdot \mathbb{1}[acc(a'_i)] \quad (2)$$

in which, $cos(a_i, a'_i)$ indicates the semantic similarity between $a_i$ and $a'_i$, $\mathbb{1}[acc(\cdot)]$ indicates whether $a'_i$ matches the ground truth from the dataset, $BLEU(\cdot)$ and $ROUGE(\cdot)$ are the Bleu and Rouge scores between answers[22, 31]. Note that, ground truth comes from the dataset and it is different from $a_i$. Details regarding the ground truth will be presented in Section 4.1. Building upon the clique assumption mentioned earlier, our overarching goal is to

$$\arg\max D_H = \frac{1}{c} \arg\max_H \sum_{k=1}^{c} d(\mathbb{A}'_k, \mathbb{A}_k)$$
$$= \frac{1}{c} \arg\max_H \sum_{k=1}^{c} d(g^{-1}(LLM(f(\mathbb{P}'_k))), LLM(\mathbb{P}_k)) \quad (3)$$
$$\text{where } H(\mathbb{P}) = \mathbb{P}'_1, \mathbb{P}'_2, \ldots, \mathbb{P}'_k, \ldots$$

In other words, we aim to find a clique function $H$, such that grouping $\mathbb{P}'_k$ inside it can achieve maximum faithfulness after the batching and dispatching process.

*3.2.2 Efficiency.* In this section, we will introduce the definition of efficiency. Let $m_a$ and $m_b$ represent two distinct prompting pipeline methods. For a given prompt $p$, we define relative cost and $c_{a \to b}$ and relative efficiency $e_{a \to b}$.

DEFINITION 1. *The relative cost $c_{a \to b}$ defines the cost ratio between $m_a$ and $m_b$.*

$$c_{a \to b} = w_{a \to b} \cdot \frac{m_a^p}{m_b^p} + \frac{LLM(m_a^p)}{LLM(m_b^p)} \quad (4)$$

*Where $m_a^p$ and $m_b^p$ signify the lengths of input tokens for methods $m_a$ and $m_b$ in response to prompt $p$, the lengths of output tokens generated by these methods are denoted as $LLM(m_a^p)$ and $LLM(m_b^p)$ respectively, $w_{a \to b}$ is the weight factor to adjust the importance between input tokens and output tokens.*

To remove the efficiency gained by incomplete outputs, we propose a weighted efficiency ratio $e_{a \to b}$.

DEFINITION 2. *The weighted relative efficiency of method $m_a$ with respect to method $m_b$, denoted as $e_{a \to b}$, is defined as follows:*

$$e_{a \to b} = \frac{t_b^p}{t_a^p} \cdot c_{a \to b} \quad (5)$$

| Symbol | Meaning |
| --- | --- |
| $p_i$ | The $i^{th}$ prompt in prompt flow. $p_i \in \mathbb{P}^m$ |
| $a_i$ | The $i^{th}$ of corresponding prompt. $a_i \in \mathbb{A}^m$ |
| $N_j$ | The $j^{th}$ user. $N_j \in \mathbb{N}^n$ |
| $C_k$ | The $k^{th}$ clique. $C_k \in \mathbb{C}^c$ |
| $\mathbb{P}$ | All prompts |
| $\mathbb{P}'$ | Prompts with the same domain. $\mathbb{P}'_k$ means the prompts are within domain $C_k$ |
| $P'$ | The batched prompt of $\mathbb{P}'$ |
| $\mathbb{A}$ | The answers corresponding to $\mathbb{P}$ |
| $\mathbb{A}'$ | The answers corresponding to $\mathbb{P}'$ |
| $A'$ | The batched answer of $\mathbb{A}'$ |
| $f(\cdot)$ | The batch function to batch prompts in $\mathbb{P}'$ to a new prompt $P'$ |
| $g(\cdot)$ | The batch function to batch prompts in $\mathbb{A}'$ to a new prompt $A'$ (We call $g^{-1}(\cdot)$ as the dispatch function to get $\mathbb{A}'$ from $A'$.) |
| $h(\cdot)$ | The clique domain function classifies a prompt $p_i$ into a specific clique domain using the function $h : \mathbb{P}^m \to \mathbb{C}^c$ |
| $H(\cdot)$ | The clique function groups prompts within the same clique, resulting in $H(\mathbb{P}) = \mathbb{P}'_1, \mathbb{P}'_2, \ldots, \mathbb{P}'_k, \ldots$ |

Table 1: Notations used in this paper

| | | | | | |
| --- | --- | --- | --- | --- | --- |
| Same input length | Running time | 1.56 | 2.04 | 20.48 | 23.29 |
| | Output length | 21 | 36 | 384 | 390 |
| Same output length | Running time | 1.17 | 1.23 | 1.33 | 1.39 |
| | Input length | 12 | 33 | 68 | 189 |

Table 2: Qualitative analysis towards running time $t_p$

Here, $t_a^p$ and $t_b^p$ denote the average execution times of methods $m_a$ and $m_b$ when processing prompt $p$. A larger value of $e_{a \to b}$ indicates greater improvement when comparing method $a$ to method $b$, suggesting that method $a$ exhibits superior performance.

THEOREM 1. *For a prompt $p$, the running time of $t_p$ for a prompt could be analyzed as:*

$$t_p = b_p + w_1 \cdot l(p) + w_2 \cdot l(LLM(p)) \quad (6)$$

in which, $b_p$ is the base running time for prompt $p$, $w_1$ and $w_2$ are coefficients for the length of input tokens and the length of output tokens. To demonstrate Equation 6, we conduct qualitative experiments as shown in Table 2. Here are the experiments we perform:

(1) Keeping a relatively small input size (10), we generate different lengths of outputs by changing the descriptive words in input prompts.
(2) Keeping a relatively small output size (10), we change different lengths of inputs by adding different lengths of background knowledge to the inputs.

This experiment demonstrates that the running time $t_p$, is directly proportional to both the length of the input and the length of the output, with each of them having a base running time $b_p$. Notably, the length of the output has a more substantial impact compared to the length of the input, as discussed in [32]. This finding further proves the importance of factoring in token lengths when calculating efficiency. Additionally, it is worth noting that the base running times $b_p$ are of similar magnitudes for different prompts. The proof is provided in the Appendix.

THEOREM 2. *For two prompts $p_1$ and $p_2$, suppose $len(p_1) \gg len(p_2)$, we still get the basic running time $\log(b_{p_1}) \approx \log(b_{p_2})$.*

PROOF. For transformer-based LLMs, suppose the prompt is with input size as $n_1$ and output size as $n_2$, the dimension and layers of the self-attention module are $d_1$, $l_1$ and the feed-forward module is $d_2$, $l_2$. Then the self-attention complexity is $O(l_1 \cdot n_1^2 \cdot d_1)$, and the feed-forward complexity is $O(l_2 \cdot d_2)$.

Since the transformer is an autoregressive model in the inference process, which means it only generates one output token per loop, the complexity becomes $n_2 \cdot (O(l_1 d_1 n_1^2 + l_2 d_2)) = O(l_1 d_1 n_2 n_1^2 + l_2 d_2 n_2)$.

If the input and output sizes are tiny, then the complexity becomes $O(l_2 d_2)$ which is the base running time. Hence, for different prompts $p_1$ and $p_2$, we have $\log(b_{p_1}) \approx \log(b_{p_2})$. □

By batching the multiple prompts $p_1, p_2, \ldots p_m$ together, we have $P'$. With the assumption of $\min d(\mathbb{A}', \mathbb{A})$, $t_{P'}$ becomes

$$\begin{aligned} t_{P'} &= b_{P'} + w_1 \cdot l(P') + w_2 \cdot l(LLM(P')) \\ &\approx b_{p_m} + w_1 \cdot \sum_{i=1}^{m} l(p_i) + w_2 \cdot \sum_{i=1}^{m} l(LLM(p_i)) \\ &= \sum_{i=1}^{m} t_{p_i} - (m-1) b_{p_m} \end{aligned} \quad (7)$$

Then we have

$$\frac{\sum_{i=1}^{m} t_{p_i}}{t_{P'}} = \frac{(m-1) b_{p_m}}{t_{P'}} + 1 \quad (8)$$

Based on the equation above, we demonstrate that increasing the size of the batching groups leads to higher relative efficiency. We also show this conclusion in Figure 7. Previous studies have often conflated the efficiency of batching with the efficiency of handling incomplete outputs. Our goal is to achieve higher efficiency while preserving the original outputs.

*3.2.3 Trade-off between faithfulness and efficiency.* To find the trade-off between faithfulness and efficiency ratio, we adopt a satisfaction model[24]. We consider which clique functions with the satisfaction model. We find the faithfulness $d(\mathbb{A}', \mathbb{A})$ and the efficiency $e_{a \to b}$ are sometimes conflicting, finding the trade-off between them becomes important.

We start by assigning different weights to faithfulness and efficiency. Then, we use the multi-objective satisfaction model[24] to achieve a trade-off between these two objectives.

## 3.3 CliqueParcel framework

Figure 2 depicts the structure of CliqueParcel. We initially employ the clique function $H$ to group users' prompts into different clusters and subsequently utilize the batch function $f(\cdot)$ to merge the prompts $\mathbb{M}'$ within these clusters into a new single prompt $M'$. After sending the new $M'$ to LLMs, we obtain the corresponding answer $A'$. Then, we dispatch the answer $A'$ to $\mathbb{A}'$ and distribute them to the users.

While the concept behind CliqueParcel is straightforward, it employs a two-stage process for batching and dispatching the original prompts. In this design, we create a batch function, denoted as $f(\cdot)$, using a simple template: *"Return the answer for each question with their corresponding numerical itemization. 1. [X] 2. [X] ..."*, where *"[X]"* represents the original prompts $p_i$. This newly constructed prompt is then inputted into the LLM to generate the corresponding answers. Subsequently, we dispatch these answers using the dispatch function $g^{-1}(\cdot)'$ along with the numerical itemization.

In our work, we systematically study different ways of clique function $H$. Given batching size $l$, and $H(\mathbb{P})_l = \mathbb{P}'_1, \mathbb{P}'_2, ...\mathbb{P}'_k, ...\mathbb{P}'_c$, in which $c = \lceil \frac{m}{l} \rceil$, $h(p_i) = C_k, \forall p_i \in \mathbb{P}'_k$. Here, $C_k$ is the clique type. We consider the following types of clique function $H$:

(1) The *concept* clique *(CC)* function $H$ groups prompts based on their topic concept. Suppose there exists a function $q$ to extract the topic concept of prompts[18], then:
$$H^* = \arg\min_H \sum_{k=1}^c \sum_{i=1}^l \left( q(\mathbb{P}'_{k_i}) - \frac{\sum_i^l q(\mathbb{P}'_{k_i})}{l} \right)^2$$
(2) The *random* clique *(RC)* function $H$ groups prompts randomly.
(3) The *semantic similarity* clique *(SSC)* function $H$ groups prompts based on their semantic similarity
$$H^* = \arg\max_H \sum_{k=1}^c \sum_{i=1}^l \sum_{j=1, j \neq i}^l cos(\mathbb{P}'_{k_i}, \mathbb{P}'_{k_j})$$
(4) The *concept plus semantic similarity* clique *(CpSC)* function $H$ initially categorizes prompts into concept-based groups and subsequently selects the smaller groups based on their semantic similarity.
(5) The *average length* clique *(ALC)* function $H$:
$$H^* = \arg\min_H \sum_{i=1}^c \left( \text{len}(\mathbb{P}'_k) - \frac{\sum (\text{len}(\mathbb{P}'_k))}{c} \right)^2$$
(6) The *maximum difference* clique *(MDC)* function $H$ groups the prompts that exhibit significant semantic differences from each other:
$$H^* = \arg\min_H \sum_{k=1}^c \sum_{i=1}^l \sum_{j=1, j \neq i}^l cos(\mathbb{P}'_{k_i}, \mathbb{P}'_{k_j})$$
(7) The *random plus average length* clique *(RpALC)* function $H$ first do random grouping then average length batching.

As a control, we also study the *separate clique function*, which is also a special case when we make $l = 1$. The framework is inspired by minibatch in training process, and for different clique function $H$, they have different performance over types of data.

## 4 EXPERIMENTS

We start by showing that CliqueParcel achieves higher performance while considering both efficiency and faithfulness trade-offs. This is necessary for efficiency-related LLM methods to restore the answers as the original one.

### 4.1 Experiment setup

*4.1.1 Benchmarks.* The datasets we use are SQuAD, HotpotQA, TREC, CSQA, GSM8K, MATH, ANLI, and MMLU. These datasets can be roughly classified into three types: reading comprehension datasets, open-source question-answering datasets, and reasoning datasets.

(1) Reading comprehension datasets are used to test models' reading comprehension abilities.
   - *SQuAD.* We utilize the Stanford Question Answering Dataset (SQuAD) [34], which is a reading comprehension dataset in our study. The questions and answers are extracted from Wikipedia articles. The dataset consists of *"context"*, *"question"*, and *"answer"*. We combine the *"context"* and *"question"* of each item to form the prompt for the language model (LLM), with the *"answer"* serving as the ground truth. SQuAD is a single-document dataset, meaning that the answer is found within one sentence in the paragraph. For the concept clique function of SQuAD, we classify them by the question type: 'what', 'when', 'where', 'who', 'why', 'how'.
   - *HotpotQA.* HotpotQA is also a Wikipedia-based dataset of question-answer pairs [39]. Similar to SQuAD, HotpotQA includes *"context"*, *"question"*, *"answer"* and one additional attribute, *"type"*. We employ the same prompt-building process for HotpotQA. In contrast to SQuAD, HotpotQA is a multi-document dataset that necessitates the use of multiple supporting documents to answer questions. In our concept clique function, we cluster the prompts by the *"type"* attribute.
(2) Open source question answering dataset owns questions that the answer could be found online.
   - *TREC.* The Text Retrieval Conference (TREC) Question Classification dataset has been widely used, primarily for information retrieval and text-based research tasks. Instead of question-answer pairs, we have chosen the TREC classification dataset to explore the impact of sophisticated concepts on the clique function $H$. We utilize the *"text"* from the TREC dataset as input prompts. In comparison to SQuAD and HotpotQA, one limitation of TREC is the absence of ground truth. We have **manually labeled** the ground truth, and this labeling has been published along with our code[1].

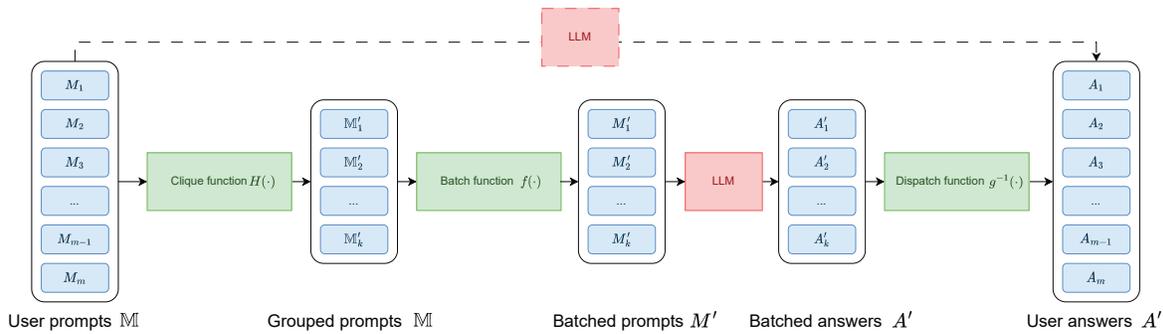

Figure 2: CliqueParcel first groups prompts from different users with a clique function $H(\cdot)$, then uses batch function $f(\cdot)$ to batch grouped prompts to a new prompt. After getting the outputs from LLM, we dispatch the outputs with dispatch function $g^{-1}(\cdot)$.

|  | separate | CC | RC | SSC | CpSC | ALC | MDC | RpALC |
| --- | --- | --- | --- | --- | --- | --- | --- | --- |
| SQuAD GPT 4 | 1.00 | 1.83 | **2.23** | 1.81 | 1.78 | 1.80 | 1.67 | 1.74 |
| SQuAD GPT 3.5 | 1.00 | 3.01 | 3.33 | **3.36** | 3.21 | 3.17 | 3.17 | 3.25 |
| HotpotQA GPT 4 | 1.00 | 1.39 | **1.86** | 1.39 | 1.41 | 1.29 | 1.36 | 1.19 |
| HotpotQA GPT 3.5 | 1.00 | 1.78 | **2.13** | 2.01 | 2.01 | 2.02 | 1.90 | 1.97 |
| TREC GPT 4 | 1.00 | 1.36 | **1.81** | 1.33 | 1.32 | 1.36 | 1.34 | 1.31 |
| TREC GPT 3.5 | 1.00 | 1.60 | **1.68** | 1.59 | 1.50 | 1.63 | 1.56 | 1.62 |
| CSQA GPT 4 | 1.00 | - | 1.90 | 1.60 | - | 1.93 | 1.73 | **1.99** |
| CSQA GPT 3.5 | 1.00 | - | **3.49** | 3.49 | - | 3.37 | 3.48 | 3.44 |
| GSM8K GPT 4 | 1.00 | - | **1.40** | 1.27 | - | 1.18 | 1.21 | 1.28 |
| GSM8K GPT 3.5 | 1.00 | - | **4.88** | 3.24 | - | 2.51 | 3.44 | 3.93 |
| MATH GPT 4 | 1.00 | - | 1.65 | 1.82 | - | 1.52 | 1.80 | **1.94** |
| MATH GPT 3.5 | 1.00 | - | 2.71 | 2.61 | - | 2.61 | 2.64 | **2.80** |
| ANLI GPT 4 | 1.00 | - | **1.03** | 0.95 | - | 1.01 | 0.80 | 0.79 |
| ANLI GPT 3.5 | 1.00 | - | **3.62** | 3.28 | - | 3.55 | 3.43 | 3.44 |
| MMLU GPT 4 | 1.00 | - | 1.77 | 1.79 | - | 1.76 | **1.82** | 1.68 |
| MMLU GPT 3.5 | 1.00 | - | 2.74 | 2.74 | - | 2.78 | 2.64 | **2.82** |

Table 3: We calculated the weighted efficiency ratio $e_{\text{CliqueParcel} \rightarrow \text{Separate}}$ across SQuAD, HotpotQA, TREC, CSQA, GSM8K, MATH, ANLI and MMLU. We test our algorithm on multiple clique functions $H$ and compare their performances.

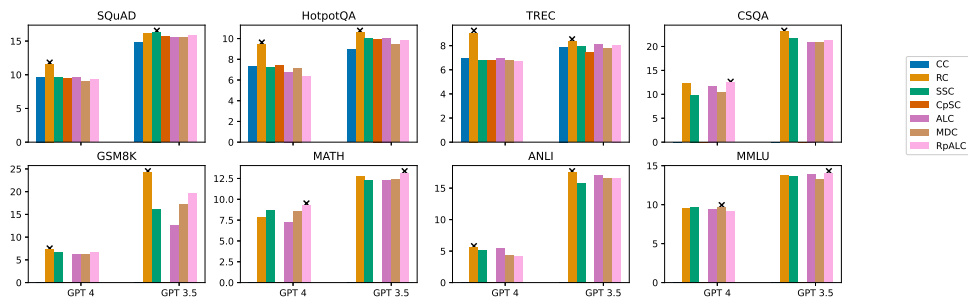

Figure 3: By employing OWA satisfaction model to efficiency and faithfulness to find a trade-off. In general, we find RC outperforms other clique functions. We label the clique function with the highest performance with a black 'x'.

---
[1]The code and the dataset are released at https://github.com/JYLEvangeline/CliqueParcel/

|  | CC | RC | SSC | CpSC | ALC | MDC | RpALC |
|---|---|---|---|---|---|---|---|
| SQuAD GPT 4 | 1.16 | 1.22 | 1.18 | 1.16 | 1.22 | 1.18 | **1.23** |
| SQuAD GPT 3.5 | 0.73 | 0.71 | 0.69 | 0.73 | **0.80** | 0.78 | 0.71 |
| HotpotQA GPT 4 | 0.81 | 0.76 | 0.78 | **0.83** | 0.78 | 0.82 | 0.77 |
| HotpotQA GPT 3.5 | **0.66** | 0.64 | 0.62 | 0.60 | 0.64 | 0.61 | 0.66 |
| TREC GPT 4 | 0.60 | 0.61 | 0.59 | **0.61** | 0.60 | 0.56 | 0.61 |
| TREC GPT 3.5 | 0.44 | 0.51 | 0.53 | 0.43 | 0.51 | 0.52 | 0.50 |
| CSQA GPT 4 | - | 2.21 | 1.78 | - | 2.10 | 1.87 | **2.24** |
| CSQA GPT 3.5 | - | **4.21** | 3.91 | - | 3.77 | 3.74 | 3.85 |
| GSM8K GPT 4 | - | **0.80** | 0.74 | - | 0.69 | 0.71 | 0.71 |
| GSM8K GPT 3.5 | - | **1.59** | 1.07 | - | 1.00 | 1.22 | 1.33 |
| MATH GPT 4 | - | 0.22 | 0.26 | - | 0.16 | 0.23 | **0.27** |
| MATH GPT 3.5 | - | 0.13 | 0.13 | - | **0.16** | 0.13 | 0.14 |
| ANLI GPT 4 | - | **0.80** | 0.72 | - | 0.77 | 0.61 | 0.60 |
| ANLI GPT 3.5 | - | **0.64** | 0.58 | - | 0.60 | 0.62 | 0.61 |
| MMLU GPT 4 | - | **1.31** | 1.30 | - | 1.28 | 1.27 | 1.31 |
| MMLU GPT 3.5 | - | **1.00** | 0.97 | - | 0.96 | 1.00 | 0.97 |

**Table 4: We calculated the faithfulness across eight datasets introduced above. We test our algorithm on multiple clique functions $H$ and compare their performances.**

(3) Reasoning datasets are another type of dataset we used, which include commonsense reasoning (CSQA), math reasoning (GSM8K, MATH), and logic reasoning (ANLI, MMLU). Since reasoning datasets lack built-in concepts, we omit CC and CpSC for them.
- **CSQA.** The Complex Sequential Question Answering (CSQA) dataset addresses factual questions through complex reasoning scenarios [35]. It presents logistical questions that require the model to engage in sequential logical reasoning processes. CSQA comprises multiple-choice questions where the model is tasked with choosing the correct answer.
- **GSM8K.** The Grade School Math 8K (GSM8K) dataset comprises basic mathematical problems typically encountered in grade school education [7]. Solutions in GSM8K primarily involve basic arithmetic operations (+, -, ×, ÷).
- **MATH.** The MATH dataset contains mathematics questions ranging from school-level to more challenging difficulties [36]. MATH offers a multitask test, and for our evaluation, we selected the arithmetic mixed task.
- **ANLI.** The Adversarial Natural Language Inference (ANLI) dataset focuses on logic reasoning tasks [28]. Given a context and a hypothesis, the model should discern the relationship between them, whether it's entailment, neutral, or contradiction.
- **MMLU.** The Massive Multitask Language Understanding (MMLU) dataset covers 57 tasks [14]. For our evaluation, we utilized the MMLU dataset's elementary mathematics task, which presents multiple-choice questions requiring the model to select the correct answer.

*4.1.2 Baselines.* For the baselines, we consider standard few-shot prompting, which we refer to as 'separate' in our experiment[3]. Examples are formatted as question-answer pairs in Figure 1 (left). Another baseline we adopt is from Cheng et al.[6], which also enhances the efficiency of standard few-shot prompting. In our comparison, we refer to the baselines proposed by Cheng et al. as SSC and MDC.

Our proposed CliqueParcel aims to improve the efficiency in few-shot prompting while keeping the answer as complete and accurate as it should be. We illustrate a sample exampler of CliqueParcel in Figure 1 (right), and propose several clique functions $H$ to explore the efficiency and faithfulness trade-offs.

In the context of our models, we begin by experimenting with various batch sizes, denoted as $l$ for each dataset. This allows us to assess efficiency performance across different datasets. Subsequently, we select the maximum batch size for each dataset to compare performance across various clique functions. This choice ensures that the models have reached their peak efficiency levels. We evaluate CliqueParcel on two large language models, *GPT 3.5 turbo* (released at 06/13/2023) [3] and *GPT 4* (released at 06/13/2023) [29]. To reduce variance, we run each of our experiments 10 times.

### 4.2 Task and efficiency performance

The efficiency performance of CliqueParcel across various clique functions, denoted as $H$, is presented in Table 3. Due to space limitations, we show each clique function with its abbreviation, which has been stated in Section 3.3.

Overall, when we consider the influence of output token length on efficiency, it becomes evident that RC (random clique function) consistently outperforms other baseline methods. Notably, when compared to the second-largest contender, typically represented by RpALC (random plus average length), RC exhibits a 10% improvement towards GPT 4 and a 4.8% improvement towards GPT-3.5, on average. This observation suggests that our algorithm's performance is particularly pronounced as the scale of the language model increases, emphasizing its efficacy on larger language models.

An interesting phenomenon emerges when we aggregate the weighted efficiency ratios across various models and datasets. It becomes apparent that the lowest performance is consistently associated with certain concepts and combinations, specifically CC, RpALC, and CpSC. This finding further substantiates why the random clique function performs better in this context: it indicates that there is no discernible pattern for the large language model (LLM) to exploit within these particular concepts and combinations.

|  | separate | CC | RC | SSC | CpSC | ALC | MDC | RpALC |
|---|---|---|---|---|---|---|---|---|
| SQuAD GPT 4 | 0.9176 | 0.9297 | 0.9570 | 0.9453 | 0.9375 | 0.9453 | 0.9414 | 0.9414 |
| SQuAD GPT 3.5 | 0.8588 | 0.8708 | 0.8661 | 0.8711 | 0.8516 | 0.8594 | 0.8516 | 0.8711 |
| HotpotQA GPT 4 | 0.9609 | 0.9727 | 0.9688 | 0.9766 | 0.9805 | 0.9688 | 0.9570 | 0.9688 |
| HotpotQA GPT 3.5 | 0.9676 | 0.9683 | 0.9688 | 0.9637 | 0.9643 | 0.9683 | 0.9839 | 0.9609 |
| TREC GPT 4 | 0.8965 | 0.8594 | 0.8301 | 0.8125 | 0.8516 | 0.8340 | 0.8125 | 0.8359 |
| TREC GPT 3.5 | 0.8906 | 0.8320 | 0.8184 | 0.8457 | 0.8262 | 0.8398 | 0.8438 | 0.8418 |
| CSQA GPT 4 | 0.8906 | - | 0.9023 | 0.8828 | - | 0.8828 | 0.8750 | 0.8828 |
| CSQA GPT 3.5 | 0.7656 | - | 0.7812 | 0.7734 | - | 0.7617 | 0.7734 | 0.7461 |
| GSM8K GPT 4 | 0.9336 | - | 0.9961 | 0.9648 | - | 0.9258 | 0.9648 | 0.9297 |
| GSM8K GPT 3.5 | 0.8633 | - | 0.7344 | 0.7383 | - | 0.7500 | 0.7227 | 0.6641 |
| MATH GPT 4 | 0.1719 | - | 0.1797 | 0.1758 | - | 0.1328 | 0.1602 | 0.1914 |
| MATH GPT 3.5 | 0.2461 | - | 0.1133 | 0.1133 | - | 0.1367 | 0.0781 | 0.0977 |
| ANLI GPT 4 | 1.0000 | - | 1.0000 | 1.0000 | - | 1.0000 | 1.0000 | 1.0000 |
| ANLI GPT 3.5 | 1.0000 | - | 1.0000 | 1.0000 | - | 1.0000 | 1.0000 | 1.0000 |
| MMLU GPT 4 | 0.9726 | - | 0.9726 | 0.9686 | - | 0.9628 | 0.9706 | 0.9686 |
| MMLU GPT 3.5 | 0.9549 | - | 0.9216 | 0.9157 | - | 0.9138 | 0.9294 | 0.9157 |

**Table 5: We calculated the accuracy across 8 datasets. For the ground truth of TREC, we manually label it and post it together with our code, in the GitHub link we provided above. The ground truth for other datasets is derived directly from the dataset itself. We mark the clique functions with at least a 2% accuracy drop in red.**

We compare our results to the baselines. On average, we achieve a 207.25% improvement (GPT 3.5) and a 70.63% improvement (GPT 4) in efficiency compared to the original method. For the other baselines [6], which we refer to as SSC (Semantic Similarity Clique function) and MDC (Maximum Difference Clique function) in our table, we achieve average improvements of 12.23% and 2.11%, respectively.

### 4.3 Efficiency and faithfulness trade-off

As we stated above, while RC performs better in the efficiency part, we want to double-check the performance of faithfulness, since the outputs should not be a *discounted* one. In Table 4, we show the faithfulness of clique functions across all datasets and models. Different from the weighted efficiency ratio, RC is no longer the top performer in faithfulness.

To strike a balance between faithfulness and efficiency, we construct a satisfaction model that combines these two factors. A higher weight assigned to one indicates a stronger preference for that factor over the other. This trade-off analysis offers a versatile solution, catering to diverse user interests in either faithfulness or efficiency, depending on their priorities.

In this paper, we employ the multi-objective satisfaction models [24] to perform a trade-off analysis. Initially, we assign varying weights to our two objectives: efficiency and faithfulness. Subsequently, we apply the Ordered Weighted Averaging (OWA) method to these two objectives. Although the performance of faithfulness varies across datasets and models when different clique functions are applied, after conducting the trade-off analysis, RC remains the top-performing method, which is shown in Figure 3. Due to page limitations, we include the full results in Figure 6 in the Appendix.

The accuracy results are shown in Table 5. Generally, the change in accuracy is relatively small compared to the improvements in efficiency. For reading comprehension datasets such as HotpotQA and SQuAD, and most reasoning datasets (CSQA, ANLI, GSM8K, MMLU), our algorithm achieves no worse accuracy on nearly all clique functions. However, for harder tasks, such as open-source QA datasets like TREC, our accuracy decreases by 4.5%. Additionally, when the model itself performs worse on specific datasets (MATH on GPT 4 and GPT 3.5, GSM8K and MMLU on GPT 4), there is a loss of accuracy.

## 5 CONCLUSION

In this paper, we address the *"discounted output"* problem. To tackle this issue, we introduce new efficiency and faithfulness measurements as two trade-off objectives for optimization. We present CliqueParcel, a framework designed to improve efficiency while maintaining output faithfulness. We evaluate CliqueParcel on eight datasets. Compared to the default framework in LLMs, CliqueParcel achieves a 138.93% improvement. Additionally, we conduct experiments to explore the relationship between batch size and efficiency, and we propose a practical parameter selection strategy for various tasks. Our work contributes to the field of LLM inference efficiency by introducing a framework that considers both efficiency and faithfulness. CliqueParcel can be easily integrated into pre-trained LLMs, and we provide multiple clique functions for different scenarios, which are helpful for various applications.

# A APPENDIX
## A.1 The Dispersion Of Length

In Figure 4 and 5b, we show the length distribution across all 8 datasets. In general, SQuAD and HotpotQA, which are reading comprehension datasets, have longer input prompts, given the necessity for context. Among reasoning datasets, ANLI, which is also context-dependent, features relatively long input prompts. From the perspective of length dispersion, the greatest differences occur in SQuAD and HotpotQA, followed by TREC, MMLU, and GSM8K. The least variation in length is observed in CSQA, MATH, and ANLI.

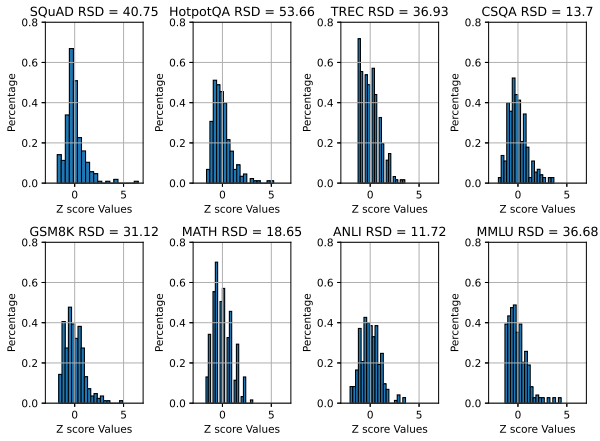

**Figure 4: This figure provides an overview of length dispersion. The title indicates the Relative Standard Deviation of lengths, while the figure illustrates the distribution of Z-scores.**

## A.2 Qualitative analysis for Table 2

Figure 5a provides a qualitative analysis for Equation 6, also an auxiliary view for the conclusion in Table 2. It proves $w_2 \gg w_1$, where $w_1$ and $w_2$ are coefficients for the length of input tokens and the length of output tokens.

## A.3 Full results for efficiency and faithfulness trade-off

We apply OWA satisfaction model to employ the trade-off between efficiency and faithfulness[24]. Before we apply the model, we assign different weights for the objectives we state above to see the performance. Figure 6 shows the performance of each clique functions. RC performs better when different weights applied.

## A.4 Batching size over running time

Figure 7 and Figure 8 provide a discussion of the relationship between size and efficiency, as well as size and running time. In this analysis, we focus on comparing the most basic clique sequence to provide experimental proof for Equation 8.

In Figure 7, we can clearly observe that as the batching size ($l$) increases, there is a significant reduction in running time. This implies that larger batching sizes can lead to time savings during processing. However, Figure 8 reveals that the efficiency gain isn't solely due to batching size. Some of the efficiency ratio improvement is attributed to smaller output token lengths.

As the batching size grows larger, the output ratio also increases, indicating that the model tends to produce more concise or 'discounted' outputs. For datasets with relatively short answers and prompts (such as TREC and MATH), compared to other datasets, there comes a point where the output ratio stabilizes, meaning that further increases in batching size do not significantly shorten the output. Conversely, reading comprehension datasets like SQuAD and HotpotQA, which are context-based question-answering datasets, exhibit lower output ratios.

In conclusion, the optimal batching size strategy depends on the nature of the task. For tasks with longer input prompts, increasing the batching size is advantageous. However, for tasks with shorter input prompts, the choice of batching size becomes a more nuanced consideration, involving user preferences and trade-offs.

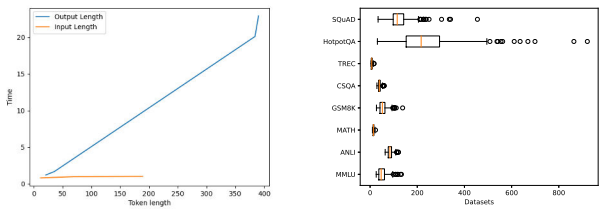

(a) Qualitive analysis for Equation 6.

(b) Box plot for length distribution.

**Figure 5: Figure 5a provides a supplementary view to Table 2. When the token length changes, the sensitivity of running time towards output length is much more than the input length, which further proves $w_2 \gg w_1$ in Theorem 1. Figure 5b describes the distribution of length.**

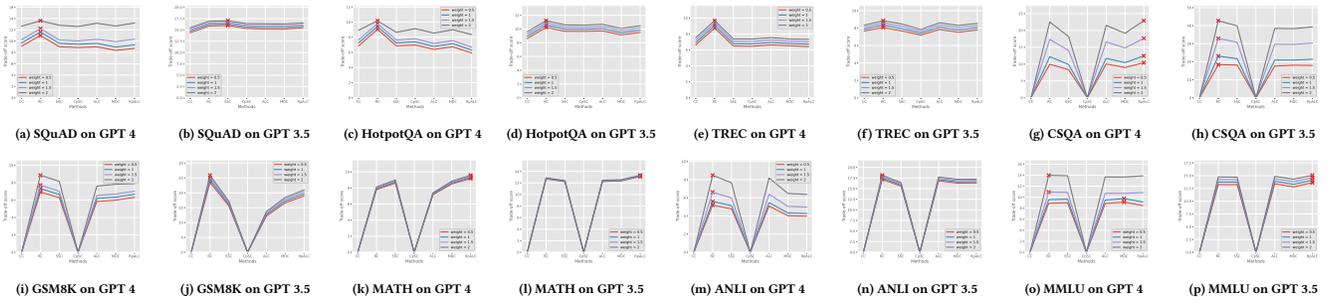

Figure 6: Trade-off between completeness and effectiveness with different weights

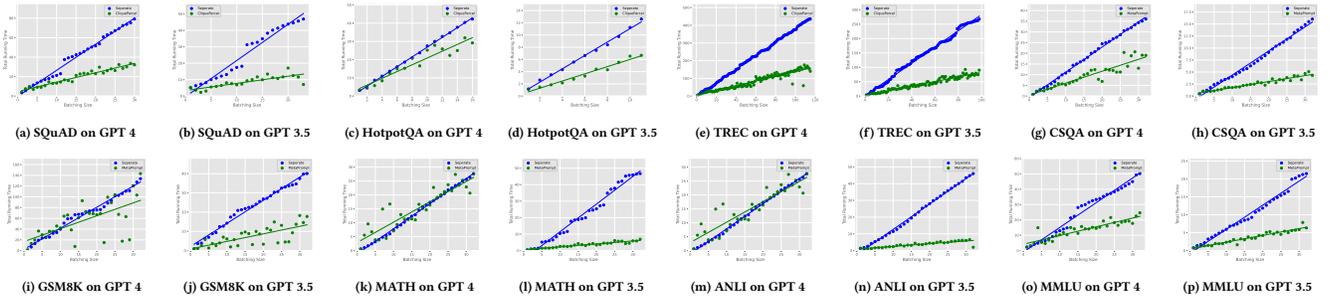

Figure 7: Running time from different batching size

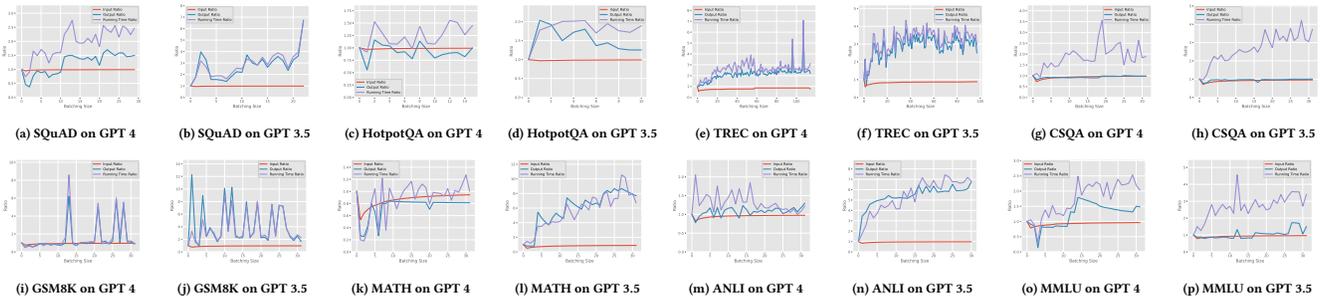

Figure 8: Gain/Ratio from different batching size